\documentclass[fleqn]{article}
\usepackage{spconf,amsmath,graphicx}
\usepackage{amsmath, amssymb}
\usepackage{float, booktabs}
\usepackage{multirow}
\usepackage{url}

\title{Guided Context Gating: Learning to leverage salient lesions in retinal fundus images}
\name{Teja Krishna Cherukuri$^{*}$, Nagur Shareef Shaik$^{*}$, Dong Hye Ye$^{*}$
}
\address{$^*$Department of Computer Science, Georgia State University, Atlanta, GA, United States}

\begin{document}
%
\maketitle
\begin{abstract}
Effectively representing medical images, especially retinal images, presents a considerable challenge due to variations in appearance, size, and contextual information of pathological signs called lesions. Precise discrimination of these lesions is crucial for diagnosing vision-threatening issues such as diabetic retinopathy. While visual attention-based neural networks have been introduced to learn spatial context and channel correlations from retinal images, they often fall short in capturing localized lesion context. Addressing this limitation, we propose a novel attention mechanism called Guided Context Gating, an unique approach that integrates Context Formulation, Channel Correlation, and Guided Gating to learn global context, spatial correlations, and localized lesion context. Our qualitative evaluation against existing attention mechanisms emphasize the superiority of Guided Context Gating in terms of explainability. Notably, experiments on the Zenodo-DR-7 dataset reveal a substantial 2.63\% accuracy boost over advanced attention mechanisms \& an impressive 6.53\% improvement over the state-of-the-art Vision Transformer for assessing the severity grade of retinopathy, even with imbalanced and limited training samples for each class.
\end{abstract}


\section{Introduction}
\label{sec:introduction}
\vspace{-2mm}
Diabetic Retinopathy is a prominent visual impairment that can lead to permanent vision loss in long-term diabetics \cite{janghorbani2000incidence}. It is characterized by accelerating vascular disruptions in the retina due to chronic hyperglycemia. The progressive nature of this disorder can result in the formation of various pathological signs called lesions \cite{shaik2022hinge}. Based on the growth of these lesions, retinopathy can be categorized as non-proliferative (NPDR) and proliferative retinopathy (PDR). NPDR includes Mild, Moderate, Severe, and Very Severe severity grades, while PDR has an additional advanced severity grade \cite{shaik2021lesion}. According to the World Health Organization (WHO), it is estimated that these diseases can affect around 500 million individuals worldwide. This number can increase even further if necessary care is not taken by ophthalmologists to identify this disease at early stages and treat accordingly,  highlighting the need for automating the diagnosis process. \cite{zheng2012worldwide}.

\begin{figure}[t]
    \centering
        \includegraphics[width=0.4\textwidth]{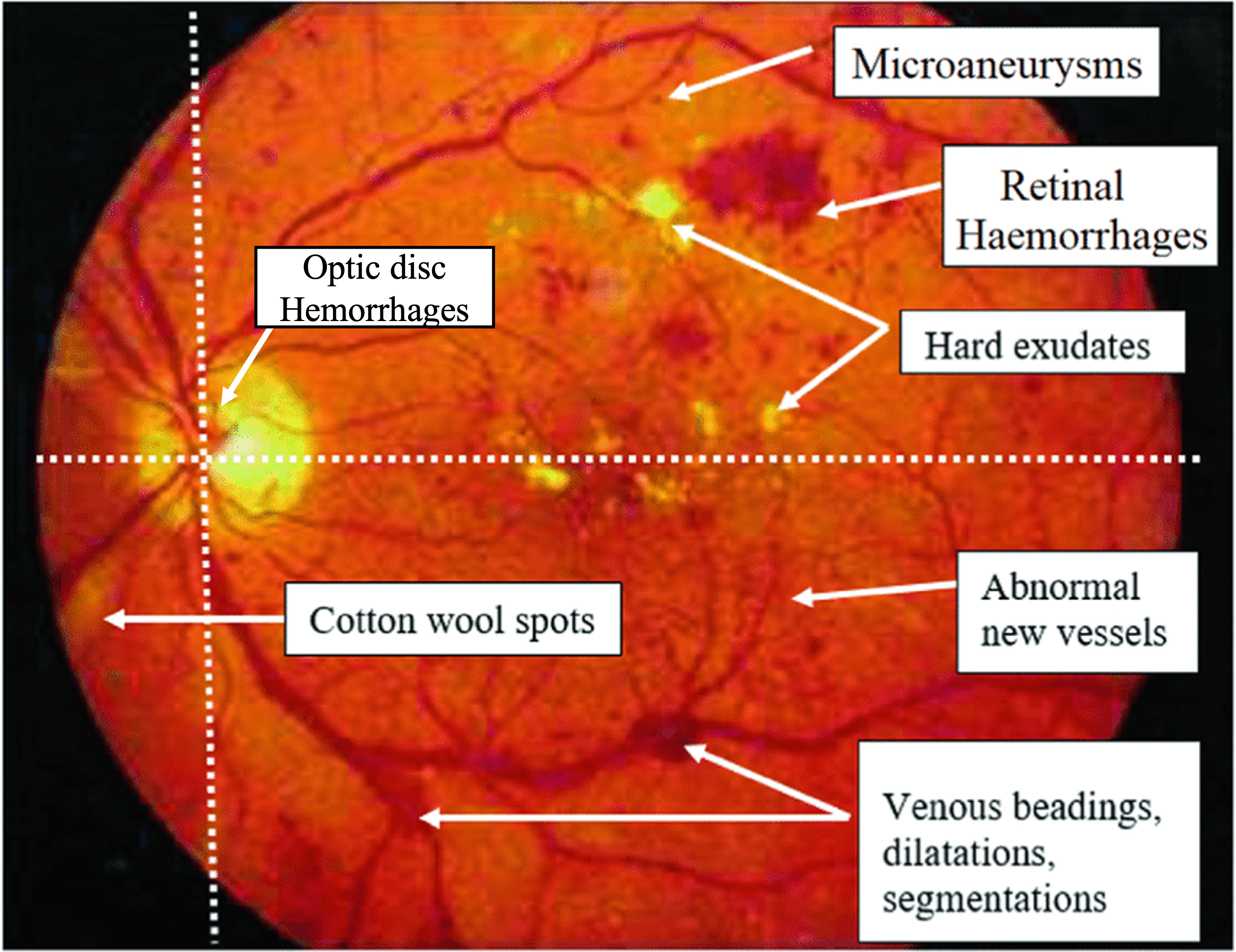}
        \vspace{-4mm}
        \caption{Retinal fundus image highlighting various lesions}
    \label{fig:lesions}
    \vspace{-6mm}
\end{figure}

Figure \ref{fig:lesions} illustrates a retinal scan with annotations highlighting various types of lesions, including micro-aneurysms, hemorrhages, exudates, and cotton wool spots. Discriminating among these lesions poses a challenge due to variations in color, shape, size, and location, making accurate identification and differentiation a complex task in the diagnosis process. Small localized bleedings at the optic nerve head, referred to as Optic disc hemorrhages, present additional layer of complexity in discrimination due to their micro size, subtle appearance, and a potential overlap with the disc. The similarity between optic disc and exudates further elevates the risk of incorrect clinical decisions, emphasizing the need for advanced representation learning techniques for accurate discrimination in the diagnosis process.

\begin{figure*}[!t]
    \centering
    \centerline{\includegraphics[width=0.895\textwidth]{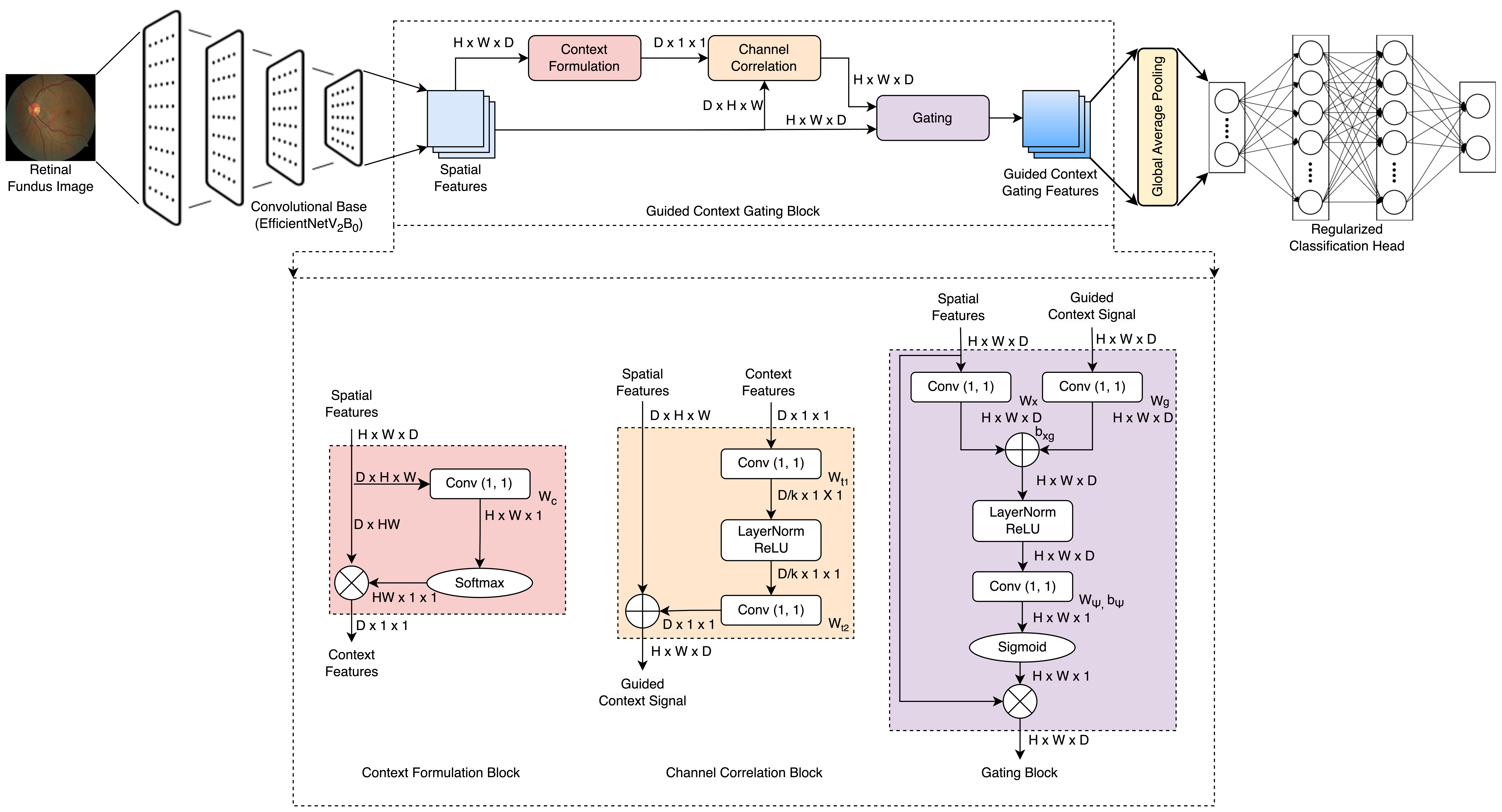}}
    \caption{Architecture of proposed Guided Context Gating Network that formulates context from convolutional features and employs it as a guiding signal for computing lesion-specific localized context, by retaining both spatial context and channel correlations; \textbf{Context Formulation} -- selectively focuses on relevant features in the initial spatial representations and computes global context information; \textbf{Channel Correlation} -- processes the computed context information \& capture channel-wise correlations; \textbf{Guided Gating} -- utilizes context features to compute lesion contextual attention representations;}
    \label{fig:architecture}
    \vspace{-4mm}
\end{figure*}

In recent years, deep convolutional neural networks (CNNs), particularly NASNet pre-trained CNN, have been utilized for learning spatial representations from medical images like retinal fundus images \cite{dondeti2020deep}. However, the lack of a dedicated attention mechanism to emphasize disease-specific features has limited the performance. A spatial attention-based method was introduced to learn texture-specific features from fundus images \cite{alahmadi2022texture}, yet it falls short in retaining category-specific information across the feature set. Channel attention-based methods, such as CANet \cite{li2019canet} and CABNet \cite{he2020cabnet}, successfully capture spatial and channel context but may miss to retain global context. Gated Attention in a composite neural network was implemented to learn lesion-specific local features from retinal images \cite{bodapati2021composite}, yet it struggled to preserve global context. Cross-lesion attention networks were designed for complex lesions based on channel-spatial convolution \cite{liu2023cross}, and Transformer-guided category-relation attention networks were aimed to enhance feature information within the class \cite{zang2024cra}. While attention-based models have shown robustness in capturing holistic features, challenges persist due to lesion variability, contrast issues, limited annotated data, class imbalance, and ethnic variations in lesion characteristics \cite{romero2024attention, madarapu2024deep}. Addressing all aforementioned challenges, we design a special module of attention called Guided Context Gating which can be embedded to any CNN architecture. This can learn to leverage the salient lesions in retinal funds images which is crucial for accurate retinopathy assessment. Guided Context Gating involves in capturing global context and identifying the importance of different spatial positions in the overall context. Significantly, it captures localized lesion context features in addition to global context representations. This holistic approach ensures a more comprehensive understanding of the retinal features, enhancing the model's ability to make accurate assessments. 
\vspace{-3mm}
\section{Methodology}
\label{sec:methods}
\vspace{-2mm}
This research aims to introduce a specialized module known as Guided Context Gating, designed to augment the capabilities of deep convolutional neural networks in effectively learning and utilizing significant lesions within retinal images, even when faced with limited training data. The comprehensive architectural details of the proposed Guided Context Gating module are illustrated in Figure \ref{fig:architecture}. The subsequent subsections provide details associated with each module involved in the proposed Guided Context Gating.

\vspace{-3mm}
\subsection{Convolutional Base}
\vspace{-2mm}
In the initial stage of our proposed framework, we focus on learning representations from retinal scan images. Traditionally, Deep CNN, with convolution and pooling operations, was employed for spatial representation acquisition from images of varied modalities. However, a drawback is the need for large amounts of labeled data for training, which is often challenging in medical image analysis. To overcome this, pre-trained CNN, such as EfficientNet$V_2$ models ($B_0$ to $B_7$) equipped with ImageNet weights, was introduced. EfficientNet$V_2$ is designed for scalability across depth, width, and resolution dimensions, incorporating special Squeeze \& Excitation blocks for attention. Our choice for this research is EfficientNet$V_2B_0$ as the Convolutional base, extracting spatial representations from retinal images. Passing a retinal scan image $X$ with dimensions $(512 \times 512 \times 3)$ through EfficientNet$V_2B_0$ results in spatial representations $R$ of dimensions $(16 \times 16 \times 1280)$ from the final convolutional block, serving as fundamental features of retinal images.

\vspace{-3mm}
\subsection{Guided Context Gating}
\vspace{-2mm}
The retinal representations learned from the pre-trained CNN exhibit global information, with equal priority given to each feature map. However, not all features may be sufficiently significant to aid the model in making accurate decisions. To address this limitation, attention mechanisms have been introduced in the literature, aiming to prioritize the most relevant features and enhance the performance of models trained on them. Spatial and Channel attention mechanisms have been widely employed for learning task-specific and cross-channel correlations. In our proposed model, the Squeeze and Excitation blocks of EfficientNet$V_2B_0$ serve this purpose, allowing the model to focus on channel specific features. Despite the usage of attention, model still falls short in capturing lesion-specific contextual information, such as lesion shapes, location, and spatial texture. These aspects are crucial for diagnosing retinal diseases, and their omission hinders the model's ability to make accurate assessments. Inspired from \cite{schlemper2019attention,cao2020global}, we design Guided Context Gating Attention block which can achieve this and enhance the retinal image representations with lesion context rich information. The modeling process can be structured into three distinct phases: context formulation, channel correlation, and guided gating.

\vspace{-3mm}
\subsubsection{Context Formulation}
\vspace{-2mm}
This module computes lesion context information by employing global attention pooling through a two-step process. First, a point-wise convolution is applied to the spatial representations, which captures essential features and relationships within the data. Subsequently, a softmax activation function is utilized to normalize and highlight the significance of different spatial elements. The combination of these two steps results in an attention map, where each element represents the importance of the corresponding spatial position in contributing to global context information. This attention map is then used to selectively focus on relevant features in the initial spatial representations, effectively computing lesion context information. For instance, when the spatial representations $R$ with dimensions $(H \times W \times D)$ are passed through a \textit{$Conv(1 \times 1)$} operation followed by \textit{Softmax}, an attention map $A$ of dimension $(H \times W \times 1)$ is obtained. The key benefit lies in the fact that this attention map is then multiplicatively combined with the initial feature maps, resulting in a lesion context attention map with dimensions $(D \times 1 \times 1)$. To achieve this, reshape and transformation operations are performed as necessary, facilitating efficient feature map multiplications. 
\begin{equation}
    A_c = \sum\limits_{j=1}^d \frac{e^{W_cr_j}}{\sum\limits_{m=1}^d e^{W_cr_m}}  \qquad \forall r_{j} \in R (r_1, r_2, ..., r_d)    \label{eq:gcp}
\end{equation}

Equation \ref{eq:gcp} represents the mathematical computation of lesion context information through global attention pooling of spatial representations $R$ where $W_c$ is point-wise convolution parameters and $A_c$ is context attention map.

\vspace{-3mm}
\subsubsection{Channel Correlation}
\vspace{-2mm}
In this module, channel-wise dependencies are captured through a specialized block comprising two point-wise convolution layers, complemented by Layer Normalization and Rectified Linear Unit (ReLU) activation functions. This strategic configuration is specifically designed for processing previously computed context information and effectively captures channel-wise correlations within retinal representations. The inclusion of ReLU activation introduces non-linearity to the processing of context information, enhancing the model's capacity to discern intricate dependencies among different channels. The resulting features are then adeptly aggregated with the original spatial features to incorporate the identified correlations among different channels in the spatial representations, thereby enriching the overall contextual understanding within the retinal features.

\vspace{-5mm}
\begin{equation}
    \delta = \sum\limits_{j=1}^{k} W_{t1}a_j \qquad \forall a_{j} \in A_c (a_1, a_2, ..., a_{k}) \label{eq:conv1}
\end{equation}
\vspace{-5mm}
\begin{equation}
    \theta = \text{LayerNormalization}(\Gamma(\delta)) \label{eq:rl}
\end{equation}
\vspace{-5mm}
\begin{equation}
    \hat{A_c} = \sum\limits_{j=1}^d W_{t2}\alpha_j \qquad \forall \alpha_{j} \in \theta (\alpha_1, \alpha_2, ..., \alpha_d) \label{eq:conv2}
\end{equation}
\vspace{-5mm}
\begin{equation}
    R_g = R \oplus \hat{A_c} \label{eq:agg}
\end{equation}

Equations \ref{eq:conv1} to \ref{eq:agg} elucidate the mathematical formulations of the channel correlation block. In these equations, $\Gamma$ represents ReLU activation, $W_{t1}$ and $W_{t2}$ represent the parameters of the point-wise convolution layers, while $a_j$ and $\alpha_j$ correspond to the elements of $A_c$ and $\theta$, respectively. Equation \ref{eq:agg} illustrates the aggregation process, where the original visual representations $R$ are combined with the computed context features $\hat{A_c}$ through element-wise addition ($\oplus$). This operation signifies the fusion of information from both the original visual features and the context features.

\vspace{-3mm}
\subsubsection{Guided Gating}
\vspace{-2mm}
This module is designed to enhance the model's ability to focus on lesion specific regions within the retinal representations. This is achieved by introducing a guiding mechanism that utilizes context features to compute attention coefficients. These are obtained through a series of convolutional operations, including point-wise convolutions and non-linearities  (ReLU and Sigmoid). The convolution operations create an additive attention map that emphasizes relevant features based on the combination of global and local context information. The attention coefficients, derived from the additive attention map, serve as a guiding signal which helps to selectively amplify or suppress features in the retinal representations based on their importance as determined by the attention coefficients. The combination of convolution operations, normalization, and gating ensures that the model can effectively highlight lesion-specific details, providing a mechanism for capturing both global and localized context features within the retinal images.

When the original retinal representations $R$ and the corresponding guiding context signal $R_g \in \mathbb{R}^{W \times H \times D}$ are passed to this module, lesion-specific coefficients $R_{l}$ are computed by considering their joint information through two point-wise convolution operations with parameters $W_x$, $W_g$. The guiding context signal $R_g$ is parametrically \(b_{rg}\) fused along with the retinal representations to jointly identify salient features at different scales by emphasizing relevant features based on the combination of global and local context information. These coefficients are further subjected to Rectified Linear Unit (ReLU) and Layer Normalization to introduce non-linearity and normalize the values, respectively. The inclusion of ReLU activation brings non-linearity to the processing coefficients, allowing the model to capture complex relationships and intricate dependencies among different features. By normalizing these coefficients across the channel dimension, Layer Normalization mitigates the internal covariate shift during training, leading to more robust and faster convergence. The gating coefficients $R_\psi$ is formulated using an additive attention mechanism, specifically utilizing a sigmoid activation function by employing a set of parameters ($W_{\psi}$, $b_{\psi}$).

\vspace{-5mm}
\begin{equation}
    R_{l,i} = \Gamma(W_x r_i + W_g g_i + b_{xg}) \qquad \forall r_{i} \in R, g_i \in R_g \label{eq:rli}
\end{equation}
\vspace{-5mm}
\begin{equation}
    R_{\psi} = \psi R_{l} + b_{\psi} \label{eq:r_att}
\end{equation}
\vspace{-5mm}
\begin{equation}
    R_{att} = \sigma(R_\psi, W_\psi) \label{eq:gcg}
\end{equation}

\noindent where $\Gamma$ is ReLU activation, element-wise non-linearity, $\psi$ is parameter vector for computing $R_{\psi}$ and $\sigma(x)$ is sigmoid activation, a normalization function to restrict the attention coefficient range to $[0, 1]$. These guided gating attention coefficients are multiplied with original retinal features to compute salient lesion representations from retinal images.

\vspace{-3mm}
\subsection{Regularized Classification Head}
\vspace{-2mm}
The Regularized Classification Head comprises two ReLU-activated fully connected dense layers with 512 and 256 neurons, respectively. Each layer integrates a batch normalization layer followed by a dropout layer with a 0.3 rate, mitigating overfitting. Moreover, $L_1$ regularization is applied to the weights, and $L_1L_2$ regularization is imposed on biases during training. These modules are arranged and trained end-to-end for predictions. Batch Normalization, Dropout, and parameter regularization layers are stacked together, culminating with a softmax layer for predicting class labels. These regularization techniques collectively enhance the model's robustness, enabling it to generate meaningful predictions with acceptable precision. The model employs Gradient Centralization \cite{yong2020gradient} during training, introducing a novel optimization technique that standardizes network activation values through Z-score standardization of weight vectors. This process can be expressed as $\mathbf{W} = \frac{W - \mu}{\sigma}$, where $W$ is the weight vector, and $\mu, \sigma$ denote the mean and standard deviation of $W$. Unlike traditional parameter operations, Gradient Centralization operates directly on gradient values, ensuring a zero mean for gradient vectors. This boosts training speed and efficiency, generalizes weight and output feature space, and improves the overall generalizability of proposed model.
\vspace{-4mm}

\section{Experiments}
\label{sec:experiments}

\vspace{-3mm}
\subsection{Dataset}
\vspace{-3mm}
\label{ssec:dataset}
The effectiveness of the proposed approach is evaluated using the Zenodo-DR-7 benchmark dataset, which consists of 757 color fundus images obtained from the Department of Ophthalmology at Hospital de Clínicas, Universidad Nacional de Asunción, Paraguay \cite{benitez2021dataset}. These retinal images were captured using the Zeiss Visucam 500 camera following a clinical procedure. Expert ophthalmologists have meticulously classified the dataset, enhancing its value for the detection of Non-Proliferative Diabetic Retinopathy (NPDR) and Proliferative Diabetic Retinopathy (PDR). The classifications include various labels such as Normal - $C_0$ (187), Mild NPDR - $C_1$(4), Moderate NPDR - $C_2$(80), Severe NPDR - $C_3$(176), Very Severe NPDR - $C_4$(108), Proliferative DR - $C_5$(88), and Advanced PDR-$C_6$(114). The dataset is divided into specific training (605) and test (152) splits. We extended our experiments to the IDRiD \cite{porwal2018indian}, Messidor-2 \cite{abramoff2013automated}, and APTOS-2019 \cite{shaik2022hinge} datasets, encompassing five severity grades $(C_0, C_1, C_2, C_3, C_5)$. Detailed information about these datasets is available in the provided references.

\vspace{-3mm}
\subsection{Experimental Setup}
\vspace{-3mm}
The proposed model was implemented using the Keras and TensorFlow libraries and experimental studies were conducted on a platform equipped with an Nvidia P100 GPU featuring 16GB Memory, a 1.32GHz clock, and supporting 9.3 TFLOPS. To optimize the model, various hyperparameters values were explored, including learning rates from 0.0001 to 0.01, dropout rates between 0.1 and 0.5, and regularization rates from 0.001 to 0.1. We selected hyperparameter values to enhance the model's performance, setting the initial learning rate to 0.0001, a batch size of 32, and training over 100 epochs. Weight regularization using L2 with a coefficient of 0.005 and bias regularization using L1\_L2 with the same coefficient were applied. The activation function for the multi-layer perceptron was ReLU, and the output activation function was Softmax. The optimizer used was RMSProp, and the loss function employed was categorical cross-entropy. During all experiments, we followed the standard hold-out-set validation strategy for saving the model checkpoint.

\vspace{-3mm}
\subsection{Quantitative Evaluation}
\vspace{-3mm}
\label{ssec:quant_evaln}

\begin{table*}[!ht]
    \centering
    \caption{Comparative Performance Analysis of Guided Context Gating and Attention based methods on Zenodo-DR-7}
    \label{tab:attention_metrics}
    \begin{tabular}{lcccccc}
        \toprule
        Approach & Accuracy (\%) & Precision (\%) & Recall (\%) & $F_1$ Score (\%) & Kappa (\%) & AUC (\%) \\
        \midrule
        No Attention & 82.89 & 68.47 & 68.18 & 68.06 & 83.27 & 96.90 \\
        Spatial Attention \cite{romero2024attention} & 86.18 & 85.92 & 85.07 & 85.40 & 83.07 & 97.27 \\
        Channel Attention \cite{he2020cabnet} & 86.84 & 86.91 & 86.39 & 86.52 & 83.79 & 96.90 \\
        Global Context Attention \cite{cao2020global} & 86.18 & 87.93 & 86.37 & 86.17 & 83.02 & 97.57 \\
        Gated Attention \cite{bodapati2021composite} & 87.50 & 87.28 & 86.84 & 86.92 & 84.84 & 96.85 \\
        \textbf{Guided Context Gating (Ours)} & \textbf{90.13} & \textbf{88.78} & \textbf{88.40} & \textbf{88.49} & \textbf{93.07} & \textbf{97.68} \\
        \bottomrule
    \end{tabular}
    \vspace{-4mm}
\end{table*}

\begin{table}[!ht]
    \centering
    \caption{Classification Report of Guided Context Gating Network for Retinopathy Classification on Zenodo-DR-7}
    \label{tab:dr_metrics}
    \setlength{\tabcolsep}{5pt}
    \begin{tabular}{cccccccc}
        \toprule
        Class & Acc & Prec & Rec & F1 & AUC & \#Test \\
        \midrule
        $C_0$ & 95.65 & 91.67 & 95.65 & 93.62 & 99.43 & 23 \\
        $C_1$ & 100.0 & 100.0 & 100.0 & 100.0 & 100.0 & 01 \\
        $C_2$ & 75.00 & 80.00 & 75.00 & 77.42 & 92.97 & 16 \\
        $C_3$ & 100.0 & 97.37 & 100.0 & 98.67 & 100 & 37 \\
        $C_4$ & 55.56 & 66.67 & 55.56 & 60.61 & 92.68 & 18 \\
        $C_5$ & 97.14 & 94.44 & 97.14 & 95.77 & 99.56 & 35 \\
        $C_6$ & 95.45 & 91.3 & 95.45 & 93.33 & 99.09 & 22 \\
        Macro & 90.13 & 88.78 & 88.4 & 88.49 & 97.80 & 152 \\
        Weighted & - & 89.51 & 90.13 & 89.73 & 98.64 & - \\
        \bottomrule
    \end{tabular}
    \vspace{-3mm}
\end{table}

Table \ref{tab:attention_metrics} presents a comprehensive performance comparison of the proposed Guided Context Gating approach with existing attention mechanisms. Our approach achieves superior results across various metrics. Specifically, the Guided Context Gating outperforms other attention mechanisms in terms of accuracy (Acc), precision (Prec), recall (Rec), $F_1$ score ($F_1$), kappa, and AUC. Notably, it achieves a remarkable accuracy of 90.13\%, showcasing its effectiveness in capturing relevant contextual information. Precision, recall, and $F_1$ score metrics demonstrate the balanced and robust performance of the proposed approach, ensuring both high accuracy and sensitivity in identifying salient features. The kappa statistic reflects substantial agreement beyond chance, emphasizing the reliability of our approach. Moreover, the AUC metric indicates the model's strong discriminatory power in distinguishing relevant features. These results collectively highlight the superior performance of our Guided Context Gating approach compared to alternative attention mechanisms.

Table \ref{tab:dr_metrics} presents the classification report for Diabetic Retinopathy (DR) severity using the proposed Guided Context Gating Network. Despite the dataset's imbalanced nature, especially in low-sampled classes, our approach demonstrates superior performance in accurately classifying these challenging instances, providing a robust solution for DR severity classification. The model exhibits exceptional accuracy, precision, recall, and $F_1$ score across all DR severity categories, excelling in both identifying normal conditions and detecting severe pathological states, particularly in Normal and Proliferative DR cases. Macro-average metrics emphasize balanced performance, and weighted average metrics underscore the model's robustness, considering varying class distributions. The Guided Context Gating Network's robust performance in low-sampled classes is crucial in medical applications, offering reliability in providing valuable insights across the entire spectrum of DR severity. 

\begin{table}[!ht]
    \centering
    \caption{Comparative Study of Guided Context Gating and Vision Transformer \cite{dosovitskiy2020image} across different datasets}
    \label{tab:comparision_study}
    \setlength{\tabcolsep}{4pt} 
    \begin{tabular}{lcccccc}
        \toprule
        Dataset & Model & Acc & Prec & Rec & F1 & AUC \\
        \midrule
        \multirow{2}{*}{Zenodo-DR-7} & GCG & 90.13 & 88.78 & 88.40 & 88.49 & 97.68 \\
         & ViT & 84.61 & 83.62 & 83.17 & 83.19 & 96.17 \\
        \cmidrule(lr){1-7}
        \multirow{2}{*}{IDRiD} & GCG & 72.14 & 68.24 & 69.56 & 68.34 & 87.51 \\
         & ViT & 61.17 & 50.53 & 45.91 & 46.18 & 76.26 \\
        \cmidrule(lr){1-7}
        \multirow{2}{*}{Messidor-2} & GCG & 80.23 & 80.81 & 69.02 & 73.85 & 87.9 \\
         & ViT & 76.79 & 75.98 & 57.17 & 61.47 & 88.53 \\
        \cmidrule(lr){1-7}
        \multirow{2}{*}{APTOS-2019} & GCG & 85.29 & 74.94 & 70.02 & 70.57 & 93.97 \\
         & ViT & 83.22 & 70.44 & 66.61 & 67.83 & 93.38 \\
        \bottomrule
    \end{tabular}
    \vspace{-3mm}
\end{table}

Table \ref{tab:comparision_study} illustrates the comparision study of Guided Context Gating (GCG) with the state-of-the-art Vision Transformer (ViT) \cite{dosovitskiy2020image} across the Zenodo-DR-7, IDRiD, Messidor-2, and APTOS-2019 datasets. GCG consistently outperforms ViT in all performance metrics across the datasets, highlighting its robustness and efficacy in diverse scenarios. This comparative study reveals the potential of GCG as a formidable solution for classification challenges, given its capability to handle limited, imbalanced data and provide accurate predictions even in challenging instances.

\vspace{-4mm}

\subsection{Qualitative Evaluation}
\vspace{-2mm}
\label{ssec:qual_evaln}
\begin{figure*}[!t]
    \centering
    \centerline{\includegraphics[width=0.95\textwidth]{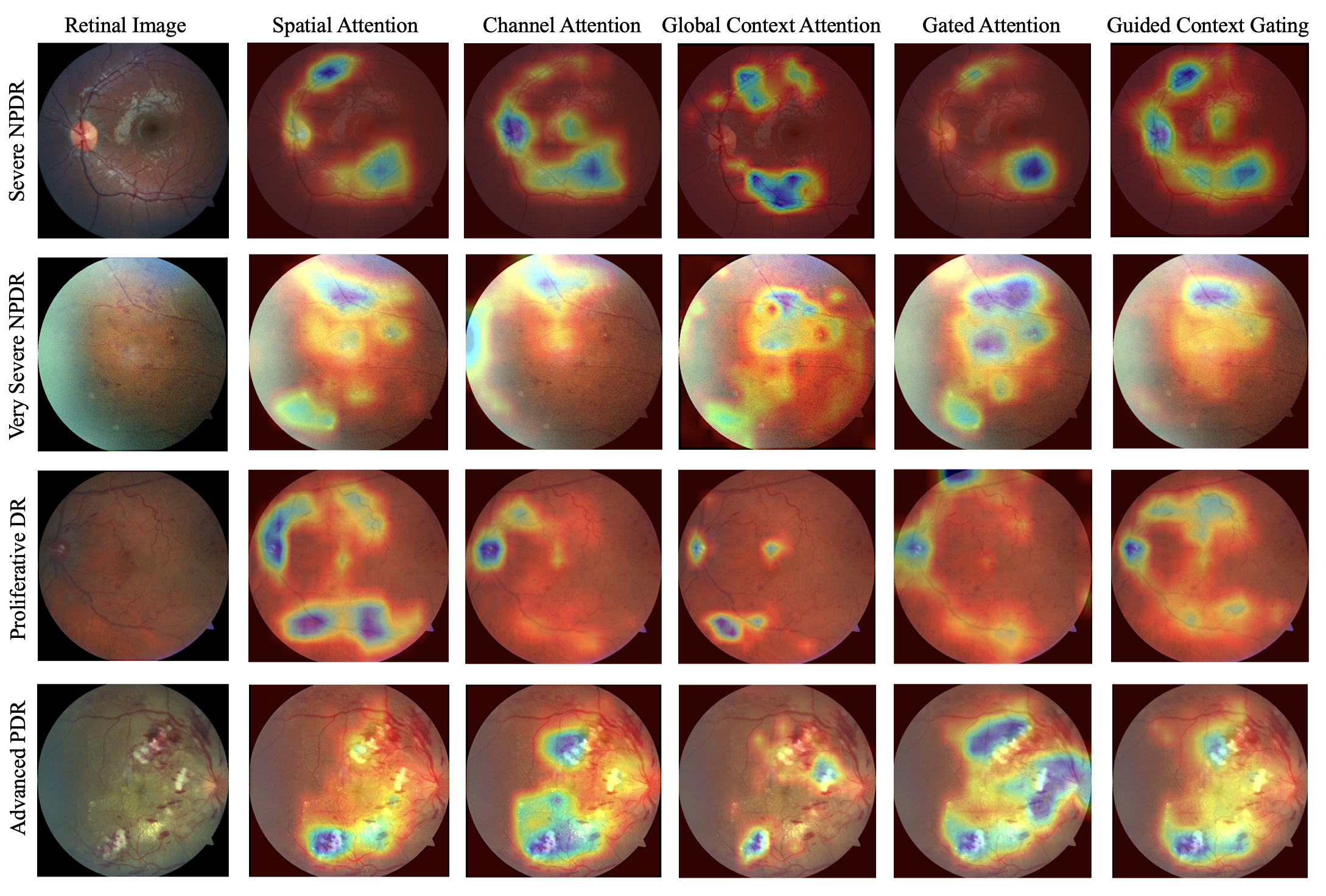}}
    \vspace{-5mm}
    \caption{Visual representation of attention maps using various strategies (spatial, channel, global context, gated, and proposed guided context gating) for severe retinopathy. Spatial attention emphasized local features but lacked broader context, and channel attention captured color and texture well but overlooked spatial context.  Global context attention risked oversimplification of lesions, and gating attention emphasized structures but occasionally highlighted unnecessary lesions. Our proposed attention highlights lesion-specific details, combining global and localized context features; dark blue indicates higher attention region.}
    \label{fig:gradcampp}
    \vspace{-3mm}
\end{figure*}

The significance of Guided Context Gating in discriminating retinopathy lies in its exceptional ability to leverage the most significant lesions from retinal images, enhancing both explainability and interpretability. As illustrated in Figure \ref{fig:gradcampp}, which depicts attention maps using different attention strategies including spatial, channel, global context, gated, and proposed Guided Context Gating, our approach stands out in highlighting the most relevant regions associated with higher severity grades of retinopathy. 

Our study revealed that spatial attention mechanisms effectively emphasized local features such as blood vessels and lesions, providing valuable insights into retinal structure. However, a notable drawback was their inability to consider broader contextual understanding, as demonstrated by their failure to capture lesions in the Advanced PDR stage. Conversely, channel attention adeptly highlighted color intensity and texture variations, capturing details like blood vessel thickness and lesion density. Yet, this focus led to the oversight of crucial spatial context, as evident in all channel attention maps. Global context attention successfully captured broader spatial relationships and contextual information, contributing to a holistic understanding of the retinal landscape. However, there was a risk of oversimplifying lesions across all severity grades, potentially undermining their significance, especially in PDR and Advanced PDR stages. Additionally, while gating attention effectively emphasized anatomical structures, concerns were raised about its tendency to highlight unnecessary retinal lesions, leading to potential misinterpretation in clinical contexts.

Guided Context Gating excels in providing clear and focused attention on lesions specific to retinopathy, effectively eliminating the attention to unwanted retinal regions. By doing so, it ensures that the model focuses solely on the clinically relevant features, enhancing its diagnostic precision. The approach successfully captures the region of interest, highlighting the areas crucial for retinopathy assessment, such as pathological lesions and abnormalities. One of the key strengths of Guided Context Gating is its ability to discriminate between intra-similar lesions, such as exudates in proximity to the optic disc, and strictly discriminate optic disc hemorrhages. While traditional attention mechanisms might struggle to distinguish between similar features, our approach ensures precise discrimination, allowing for a more accurate assessment of the severity of retinopathy. This discrimination is essential for providing ophthalmologists with a clear and interpretable visualization of the regions contributing to the retinopathy diagnosis, aiding in clinical decision-making.

\vspace{-3mm}
\section{Conclusion}
\label{sec:conclusion}
\vspace{-2mm}
This research introduces the Guided Context Gating module which effectively extracts both global and local contextual features from retinal images, addressing limitations associated with limited training data. Through comprehensive experiments on benchmark datasets, the Guided Context Gating approach outperforms existing attention mechanisms and state-of-the-art ViT in terms of all performance metrics. Notably, it demonstrates remarkable accuracy in classifying various severity levels of diabetic retinopathy, showcasing its robustness in handling imbalanced data and providing accurate predictions even for underrepresented classes. The model's superior performance is further highlighted in its ability to discriminate intra-similar lesions, contributing to precise diagnostic assessments. Future work could involve extending the proposed approach to other medical imaging tasks and integrating additional modalities of retinal pathology. 

\bibliographystyle{IEEEbib}
\bibliography{refs}

\begin{thebibliography}{10}

\bibitem{janghorbani2000incidence}
Mohsen Janghorbani, Raymond~B Jones, and Simon~P Allison,
\newblock ``Incidence of and risk factors for proliferative retinopathy and its association with blindness among diabetes clinic attenders,''
\newblock {\em Ophthalmic Epidemiology}, vol. 7, no. 4, pp. 225--241, 2000.

\bibitem{shaik2022hinge}
Nagur~Shareef Shaik and Teja~Krishna Cherukuri,
\newblock ``Hinge attention network: A joint model for diabetic retinopathy severity grading,''
\newblock {\em Applied Intelligence}, vol. 52, no. 13, pp. 15105--15121, 2022.

\bibitem{shaik2021lesion}
Nagur~Shareef Shaik and Teja~Krishna Cherukuri,
\newblock ``Lesion-aware attention with neural support vector machine for retinopathy diagnosis,''
\newblock {\em Machine Vision and Applications}, vol. 32, no. 6, pp. 1--13, 2021.

\bibitem{zheng2012worldwide}
Yingfeng Zheng, Mingguang He, and Nathan Congdon,
\newblock ``The worldwide epidemic of diabetic retinopathy,''
\newblock {\em Indian journal of ophthalmology}, vol. 60, pp. 428, 2012.

\bibitem{dondeti2020deep}
Venkatesulu Dondeti, Jyostna~Devi Bodapati, Shaik~Nagur Shareef, and Naralasetti Veeranjaneyulu,
\newblock ``Deep convolution features in non-linear embedding space for fundus image classification.,''
\newblock {\em Rev. d'Intelligence Artif.}, vol. 34, no. 3, pp. 307--313, 2020.

\bibitem{alahmadi2022texture}
Mohammad~D Alahmadi,
\newblock ``Texture attention network for diabetic retinopathy classification,''
\newblock {\em IEEE Access}, vol. 10, pp. 55522--55532, 2022.

\bibitem{li2019canet}
Xiaomeng Li, Xiaowei Hu, Lequan Yu, Lei Zhu, Chi-Wing Fu, and Pheng-Ann Heng,
\newblock ``Canet: cross-disease attention network for joint diabetic retinopathy and diabetic macular edema grading,''
\newblock {\em IEEE transactions on medical imaging}, vol. 39, no. 5, pp. 1483--1493, 2019.

\bibitem{he2020cabnet}
Along He, Tao Li, Ning Li, Kai Wang, and Huazhu Fu,
\newblock ``Cabnet: Category attention block for imbalanced diabetic retinopathy grading,''
\newblock {\em IEEE Transactions on Medical Imaging}, vol. 40, no. 1, pp. 143--153, 2020.

\bibitem{bodapati2021composite}
Jyostna~Devi Bodapati, Nagur~Shareef Shaik, and Veeranjaneyulu Naralasetti,
\newblock ``Composite deep neural network with gated-attention mechanism for diabetic retinopathy severity classification,''
\newblock {\em Journal of Ambient Intelligence and Humanized Computing}, vol. 12, no. 10, pp. 9825--9839, 2021.

\bibitem{liu2023cross}
Xiang Liu and Wei Chi,
\newblock ``A cross-lesion attention network for accurate diabetic retinopathy grading with fundus images,''
\newblock {\em IEEE Transactions on Instrumentation and Measurement}, 2023.

\bibitem{zang2024cra}
Feng Zang and Hui Ma,
\newblock ``Cra-net: Transformer guided category-relation attention network for diabetic retinopathy grading,''
\newblock {\em Computers in Biology and Medicine}, p. 107993, 2024.

\bibitem{romero2024attention}
Roberto Romero-Ora{\'a}, Mar{\'\i}a Herrero-Tudela, Mar{\'\i}a~I L{\'o}pez, Roberto Hornero, and Mar{\'\i}a Garc{\'\i}a,
\newblock ``Attention-based deep learning framework for automatic fundus image processing to aid in diabetic retinopathy grading,''
\newblock {\em Computer Methods and Programs in Biomedicine}, vol. 249, pp. 108160, 2024.

\bibitem{madarapu2024deep}
Sandeep Madarapu, Samit Ari, and KK~Mahapatra,
\newblock ``A deep integrative approach for diabetic retinopathy classification with synergistic channel-spatial and self-attention mechanism,''
\newblock {\em Expert Systems with Applications}, p. 123523, 2024.

\bibitem{schlemper2019attention}
Jo~Schlemper, Ozan Oktay, Michiel Schaap, Mattias Heinrich, Bernhard Kainz, Ben Glocker, and Daniel Rueckert,
\newblock ``Attention gated networks: Learning to leverage salient regions in medical images,''
\newblock {\em Medical image analysis}, vol. 53, pp. 197--207, 2019.

\bibitem{cao2020global}
Yue Cao, Jiarui Xu, Stephen Lin, Fangyun Wei, and Han Hu,
\newblock ``Global context networks,''
\newblock {\em IEEE Transactions on Pattern Analysis and Machine Intelligence}, 2020.

\bibitem{yong2020gradient}
Hongwei Yong, Jianqiang Huang, Xiansheng Hua, and Lei Zhang,
\newblock ``Gradient centralization: A new optimization technique for deep neural networks,''
\newblock in {\em European Conference on Computer Vision}. Springer, 2020, pp. 635--652.

\bibitem{benitez2021dataset}
Veronica Elisa~Castillo Ben{\'\i}tez, Ingrid~Castro Matto, Julio C{\'e}sar~Mello Rom{\'a}n, Jos{\'e} Luis~V{\'a}zquez Noguera, Miguel Garc{\'\i}a-Torres, Jordan Ayala, Diego~P Pinto-Roa, Pedro~E Gardel-Sotomayor, Jacques Facon, and Sebastian~Alberto Grillo,
\newblock ``Dataset from fundus images for the study of diabetic retinopathy,''
\newblock {\em Data in brief}, vol. 36, pp. 107068, 2021.

\bibitem{porwal2018indian}
Prasanna Porwal, Samiksha Pachade, Ravi Kamble, Manesh Kokare, Girish Deshmukh, Vivek Sahasrabuddhe, and Fabrice Meriaudeau,
\newblock ``Indian diabetic retinopathy image dataset (idrid): a database for diabetic retinopathy screening research,''
\newblock {\em Data}, vol. 3, no. 3, pp. 25, 2018.

\bibitem{abramoff2013automated}
Michael~D Abr{\`a}moff, James~C Folk, Dennis~P Han, Jonathan~D Walker, David~F Williams, Stephen~R Russell, Pascale Massin, Beatrice Cochener, Philippe Gain, Li~Tang, et~al.,
\newblock ``Automated analysis of retinal images for detection of referable diabetic retinopathy,''
\newblock {\em JAMA ophthalmology}, vol. 131, no. 3, pp. 351--357, 2013.

\bibitem{dosovitskiy2020image}
Alexey Dosovitskiy, Lucas Beyer, Alexander Kolesnikov, Dirk Weissenborn, Xiaohua Zhai, Thomas Unterthiner, Mostafa Dehghani, Matthias Minderer, Georg Heigold, Sylvain Gelly, et~al.,
\newblock ``An image is worth 16x16 words: Transformers for image recognition at scale,''
\newblock {\em arXiv preprint arXiv:2010.11929}, 2020.

\end{thebibliography}

\end{document}